%% file: main.tex
\documentclass[10pt,twocolumn,letterpaper]{article}

\usepackage{cvpr}              
\usepackage{bm}

\input{preamble}
\definecolor{cvprblue}{rgb}{0.21,0.49,0.74}
\usepackage[pagebackref,breaklinks,colorlinks,allcolors=cvprblue]{hyperref}

\def\paperID{14595} 
\def\confName{CVPR}
\def\confYear{2025}

\title{Curriculum Coarse-to-Fine Selection for High-IPC Dataset Distillation}

\author{%
Yanda Chen\footnotemark[1] \quad Gongwei Chen\footnotemark[1] \quad Miao Zhang\footnotemark[2] \quad Weili Guan \quad Liqiang Nie \\
School of Computer Science and Technology, Harbin Institute of Technology, Shenzhen \\
\texttt{cydaaa30@gmail.com} \\
\texttt{\{chengongwei,zhangmiao,guanweili,nieliqiang\}@hit.edu.cn}\\
}

\begin{document}
\maketitle

\renewcommand{\thefootnote}{\fnsymbol{footnote}} 
\footnotetext[1]{Equal contribution}
\footnotetext[2]{Corresponding authors}

\input{sec/0_abstract}    
\input{sec/1_intro}
\input{sec/2_related_work}

\input{sec/3_preliminary}
\input{sec/4_method}
\input{sec/5_experiment}

\input{sec/6_conclusion}
\input{sec/7_acknowledgement}

{
    \small
    \bibliographystyle{ieeenat_fullname}
    \bibliography{main}
}
\input{sec/X_appendix}
\end{document}

%% file: preamble.tex
\usepackage{threeparttable}
\usepackage{adjustbox}
\usepackage{pifont}

\usepackage{multirow}
\usepackage{amsmath}
\usepackage{algorithm}
\usepackage{algpseudocode}
\usepackage{graphicx}
\usepackage{xcolor}
\usepackage{lineno}

\newcommand{\algopt}{CCFS\xspace}
\newcommand{\IPC}{\texttt{IPC}\xspace}

\providecommand{\cA}{\mathcal{A}}
\providecommand{\cD}{\mathcal{D}}
\providecommand{\cS}{\mathcal{S}}
\providecommand{\cT}{\mathcal{T}}

%% file: sec/0_abstract.tex
\begin{abstract}
\par\addvspace{-3mm}
\noindent Dataset distillation (DD) excels in synthesizing a small number of images per class (IPC) but struggles to maintain its effectiveness in high-IPC settings.
Recent works on dataset distillation demonstrate that combining distilled and real data can mitigate the effectiveness decay. 
However, our analysis of the combination paradigm reveals that the current one-shot and independent selection mechanism induces an incompatibility issue between distilled and real images. 
To address this issue, we introduce a novel curriculum coarse-to-fine selection (CCFS) method for efficient high-IPC dataset distillation.
CCFS employs a curriculum selection framework for real data selection, where we leverage a coarse-to-fine strategy  to select appropriate real data based on the current synthetic dataset in each curriculum.
Extensive experiments validate CCFS, surpassing the state-of-the-art by +6.6\% on CIFAR-10, +5.8\% on CIFAR-100, and +3.4\% on Tiny-ImageNet under high-IPC settings.
Notably, CCFS achieves 60.2\% test accuracy on ResNet-18 with a 20\% compression ratio of Tiny-ImageNet, closely matching full-dataset training with only 0.3\% degradation.
Code: \url{https://github.com/CYDaaa30/CCFS}.
\end{abstract}

%% file: sec/1_intro.tex
\vspace{-2mm}
\section{Introduction}
\label{sec:intro}
Dataset distillation~\cite{wang2018dataset, yu2023review} aims to condense the original training dataset into a small but powerful synthetic dataset, which can then be used to train competitive models. Current Dataset Distillation (DD) methods~\cite{wu2024robust, cazenavette2022mtt, zhao2023dm}  have shown impressive performance at extremely small scales, such as 1 or 5 IPC (images-per-class). Unfortunately, these methods become less effective as IPC increases~\cite{cui2022dcbench, zhou2023dq, guo2024lossless}, sometimes even underperforming random sample selection.

Recent studies~\cite{guo2024lossless,lee2024selmatch} investigated this phenomenon and attributed it to a key issue in dataset distillation: current approaches tend to distill simple and general features into synthetic images while ignoring rare and complex features. These works attempt to address this issue from perspectives of optimization and dataset construction. The former work, DATM~\cite{guo2024lossless}, attempts to leverage the trajectories from different training stages to generate synthetic images with diverse patterns. It still fails to adequately incorporate the rare features of hard samples~\cite{lee2024selmatch}. The latter work, SelMatch~\cite{lee2024selmatch},  advocates a combination-based paradigm which merges a distilled image set $\mathcal{D}_\textrm{distill}$ and a real image set $\mathcal{D}_\textrm{real}$ to construct the synthetic dataset. By complementing rare and complex features of real data, SelMatch achieves state-of-the-art performance in high-IPC situations.

\begin{figure}[t]
    \centering
        \setlength{\abovecaptionskip}{0cm}
        \setlength{\belowcaptionskip}{-5mm}
        \includegraphics[width=1\linewidth]{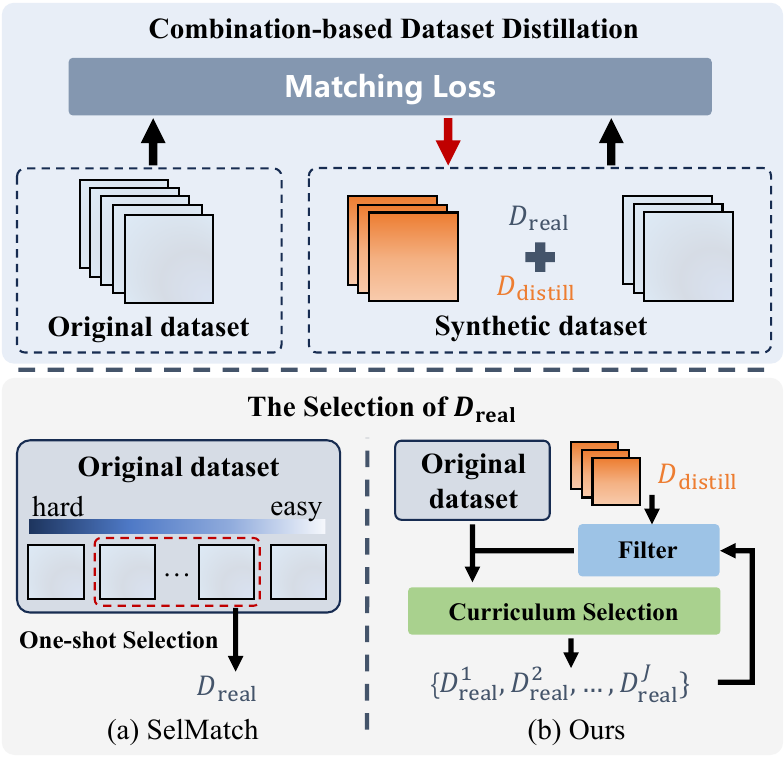}
    \caption{\textbf{Comparison of combination-based dataset distillation.}  
\textit{Top:} General paradigm. \textit{Bottom:} (a) SelMatch conducts an independent and one-shot selection of $\mathcal{D}_\textrm{real}$. (b) Our method applies curriculum selection, making $\mathcal{D}_\textrm{real}$ dependent on $\mathcal{D}_\textrm{distill}$.}
    \label{fig:contrast}
\end{figure}

A vital advantage of the combination-based paradigm is the introduction of a real image set $\mathcal{D}_\textrm{real}$. Although SelMatch shows impressive performance, we argue that its selection approach of $\mathcal{D}_\textrm{real}$ still has two shortcomings. 1) The fixed and one-shot selection of $\mathcal{D}_\textrm{real}$ is sub-optimal and may produce inappropriate real images. 2) The independence between $\mathcal{D}_\textrm{real}$ and $\mathcal{D}_\textrm{distill}$ reduces the complementary effect of $\mathcal{D}_\textrm{real}$. To verify our point, we compare SelMatch with two naive variants in Section~\ref{limiation}, ``SelMatch w/ two-shot selection'' and ``SelMatch w/ reverse selection'' which only consider a two-shot paradigm and reverse order of selection and distillation. The superiority of proposed variants reveals the underlying incompatibility issue between $\mathcal{D}_\textrm{real}$ and $\mathcal{D}_\textrm{distill}$ in SelMatch. This issue reduces the information richness of the generated synthetic dataset, distinctly impacting SelMatch's performance in high-IPC situations.

To address the incompatibility issue, we propose a novel Curriculum Coarse-to-Fine Selection (CCFS) method for high-IPC dataset distillation. CCFS aims to progressively select suitable real data based on the distilled set. We cast the selection of the real images as a curriculum learning problem. This allows us to merge the real images from easy to difficult through a series of curriculum phases, ensuring comprehensive coverage of the essential patterns. To enhance the connection between the real images and the distilled images, we devise a coarse-to-fine selection strategy that takes into account both the global  sample difficulty and the current synthetic dataset. Our selection strategy coarsely filters out correctly-classified samples, then finely chooses a subset from misclassified samples according to their difficulty scores. The coarse stage ensures that the selected samples contain the unlearned patterns of the current synthetic set. The fine stage maximizes the complementary effect of selected samples by avoiding overly complex ones.

\begin{figure*}[t]
    \centering
        \setlength{\abovecaptionskip}{-2mm}
        \setlength{\belowcaptionskip}{-5mm}
        \includegraphics[width=1\linewidth]{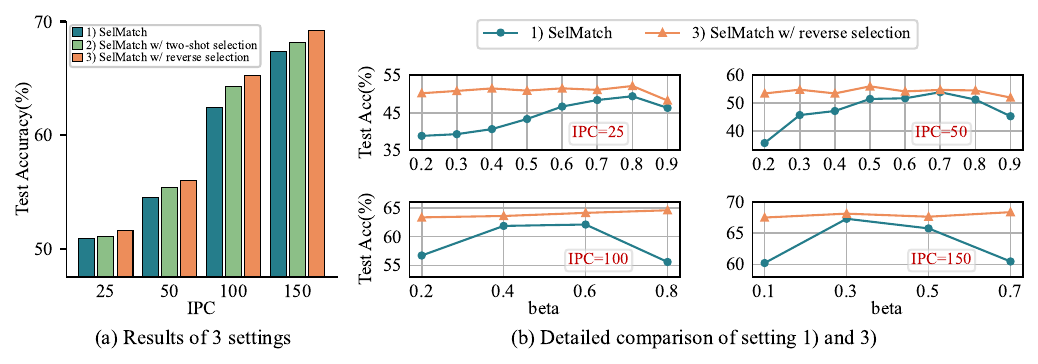}
    \caption{\textbf{Results of the analysis experiments on CIFAR-100.} (a) Top-1 accuracy of the 3 settings with IPC=25, 50, 100, 150. In each IPC, setting~\ref{group:b}, which modifies only the selection strategy of $\mathcal{D}_\textrm{real}$, outperforms setting~\ref{group:a} with the original SelMatch setup. Setting~\ref{group:c} reverses setting~\ref{group:b}’s process by first distilling $\mathcal{D}_\textrm{distill}$ and then conducting a two-shot selection to obtain $\mathcal{D}_\textrm{real}$, resulting in the best performance among the 3 groups. (b) A detailed comparison between setting~\ref{group:a} and~\ref{group:c} at various window starting point $\beta$. In all cases of $\beta$, setting~\ref{group:c} outperforms setting~\ref{group:a} and shows more stable performance fluctuations across different $\beta$.}
    \label{fig:analysis_exp}
\end{figure*}

Extensive experiments on CIFAR-10/100~\cite{krizhevsky2009cifar} and Tiny-ImageNet~\cite{le2015tiny} demonstrate that our approach consistently outperforms current state-of-the-art methods across compression ratios range of 5\%-30\%.
Specifically, we achieved top-1 test accuracies of \{92.5\%, 71.5\%,  60.2\%\} on \{CIFAR-10, CIFAR-100, Tiny-ImageNet\} with compression ratios of \{10\%, 10\%, 20\%\}, surpassing the current SOTA by \{6.6\%, 5.8\%, 3.4\%\}, separately. We record the state in each curriculum phase, including filter performance, difficulty distribution and visualization of real samples. The analyses of these states effectively illustrate the incremental growth in both performance and difficulty introduction of the synthetic dataset as the curriculum progresses, which aligns with our design principles. Our contributions can be summarized as follows:

\begin{enumerate}
    \item We advocate the combination-based paradigm for high-IPC dataset distillation, and propose a novel curriculum coarse-to-fine selection method to address the existing incompatibility issue.
    \item In each curriculum, we devise a coarse-to-fine selection strategy to yield the optimal real data by inspecting the limitations of the current synthetic dataset.
    \item Our method achieves only 0.3\% performance loss with a 20\% compression ratio on Tiny-ImageNet. The analyses indicate the selected images become more complex and difficult as the curriculum progresses.
\end{enumerate}

%% file: sec/2_related_work.tex
\section{Related Works}
\label{sec:formatting}
\vspace{-1mm}
\paragraph{Dataset Distillation.} 
Dataset distillation aims to condense the original training dataset $\mathcal{T}$ into a small but powerful synthetic set $\mathcal{S}$, which allows models to be trained efficiently and achieve performance comparable to full dataset training. Wang et al.~\cite{wang2018dataset} first proposed the concept of dataset distillation with a bi-level framework. Subsequently, a few works have sought to optimize dataset distillation from different perspectives~\cite{zhao2021DSA,bohdal2020flexible,zhang2023accelerating,he2024mdc,xu2024distill}. Meanwhile, various surrogate objectives gradually formed several significant branches, which can be summarized as kernel-based approaches~\cite{nguyen2021kip,zhou2022krr,loo2022efficient}, gradient/trajectory-based methods ~\cite{zhao2021DC,cazenavette2022mtt,guo2024lossless,du2023minimizing,liu2024dataset}, distribution-based techniques~\cite{wang2022cafe,zhao2023dm,zhao2023improved,zhang2024dance}, distilled dataset parameterization~\cite{kim2022para,deng2022remember,liu2022factor,wei2023sparse} and decoupled-optimization~\cite{yin2023sre2l,yin2023cda,sun2024rded,ma2024cudd}.

Recent studies~\cite{guo2024lossless,lee2024selmatch} have noticed the issue of dataset distillation failing when synthesizing large number of images per class (IPC). Their consensus is introducing more complex features into the synthetic dataset to alleviate its homogeneous and simplistic nature. DATM~\cite{guo2024lossless} employs a flexible trajectory matching to align expert trajectories from later training stages, aligning with the theory that models learn more complex patterns as training progresses. On the other hand, SelMatch~\cite{lee2024selmatch} emphasizes the initialization of the synthetic dataset by using a sliding window to incorporate real images of appropriate difficulty levels. It divides the synthetic dataset into two subsets, $\mathcal{D}_\textrm{distill}$ and $\mathcal{D}_\textrm{real}$, allowing for updates to $\mathcal{D}_\textrm{distill}$ while keeping $\mathcal{D}_\textrm{real}$ fixed during MTT~\cite{cazenavette2022mtt} distillation. This approach serves as the SOTA approach for high-IPC dataset distillation.

Our method CCFS is based on the concept of combining distilled data with real data to solve the failing problem in high-IPC cases. Unlike previous approaches, we design a novel curriculum framework to progressively select suitable real data for the distilled data, and conduct a more targeted selection strategy within each curriculum phase.
\vspace{-5mm}
\paragraph{Curriculum Learning in Dataset Distillation}
Curriculum learning~\cite{bengio2009curriculum,kumar2010self,lee2011learning,zhang2015self,soviany2020curriculum} is originally defined as a method for progressively training models by strategically arranging the inputting sequence of training data.  Some dataset distillation methods leverage the concept of curriculum learning. SeqMatch~\cite{du2023seqmatch} divides synthetic data into multiple subsets and sequentially optimizes them to learn high-level features. CDA~\cite{yin2023cda} implements a progressive difficulty data augmentation on the synthetic images. CUDD~\cite{ma2024cudd} employs curriculum evaluation to gradually expand the distilled dataset. In CCFS, we design a curriculum framework that gradually expands the synthetic dataset by incorporating suitable real samples. This process takes into account both the prior knowledge of sample difficulty and the limitations of the current synthetic dataset. Our curriculum framework effectively enriches the diversity of the synthetic dataset and enhances its performance.

%% file: sec/3_preliminary.tex
\section{Preliminary} \label{preliminary}

\subsection{Combination-Based Dataset Distillation }

Recent studies~\cite{cui2022dcbench,zhou2023dq,guo2024lossless} reveal that traditional dataset distillation methods tend to synthesize simple features of the original dataset. This limits its effectiveness especially when the synthetic dataset contains more images per class (IPC). To address this problem, SelMatch~\cite{lee2024selmatch} introduces a combination-based framework consisting of selection-based initialization and partial optimization.

SelMatch introduces modifications to the traditional optimization-based method in both the initialization and the updating phase. It begins by arranging the training samples of the original dataset in descending order of difficulty based on pre-calculated difficulty scores. Then it uses a sliding window of size $IPC$ (images-per-class) to select subsets in each class with a window starting point hyperparameter $\beta \in \left[ 0, 1 \right]$. It collects these selected subsets of each class as $\mathcal{D}_{\text{initial}}$ to initialize $\mathcal{D}_{\text{syn}}$. 

Once the window starting point is determined, SelMatch further partitions samples within the window according to a distillation portion hyperparameter $\alpha \in [0, 1]$. The subset $\mathcal{D}_{\textrm{real}}$ contains the harder samples of the first $(1-\alpha) \times |\mathcal{D}_{\text{syn}}|$ portion of the window and keeps unchanged during distillation. The remaining $\alpha \times |\mathcal{D}_{\text{syn}}|$ easier samples serve as the initialization set $\mathcal{D}_{\text{pre-distill}}$ for distillation to produce the subset $\mathcal{D}_{\text{distill}}$.
Both $\beta$ and $\alpha$ are optimal hyperparameters determined through a search process.

During subsequent MTT~\cite{cazenavette2022mtt} distillation, the update aims to minimize the matching loss between the entire $\mathcal{D}_{\text{syn}} = \mathcal{D}_{\textrm{real}} \cup \mathcal{D}_{\text{distill}}$ and the original dataset $\mathcal{T}$, i.e., 

\begin{equation}\label{eq3}
    \mathcal{L}\left( \mathcal{D}_{\textrm{real}} \cup \mathcal{D}_{\text{distill}}, \mathcal{T} \right).
\end{equation}

 \begin{figure*}[t]
    \centering
        \setlength{\abovecaptionskip}{0cm}
        \setlength{\belowcaptionskip}{-5mm}
        \includegraphics[width=1\linewidth]{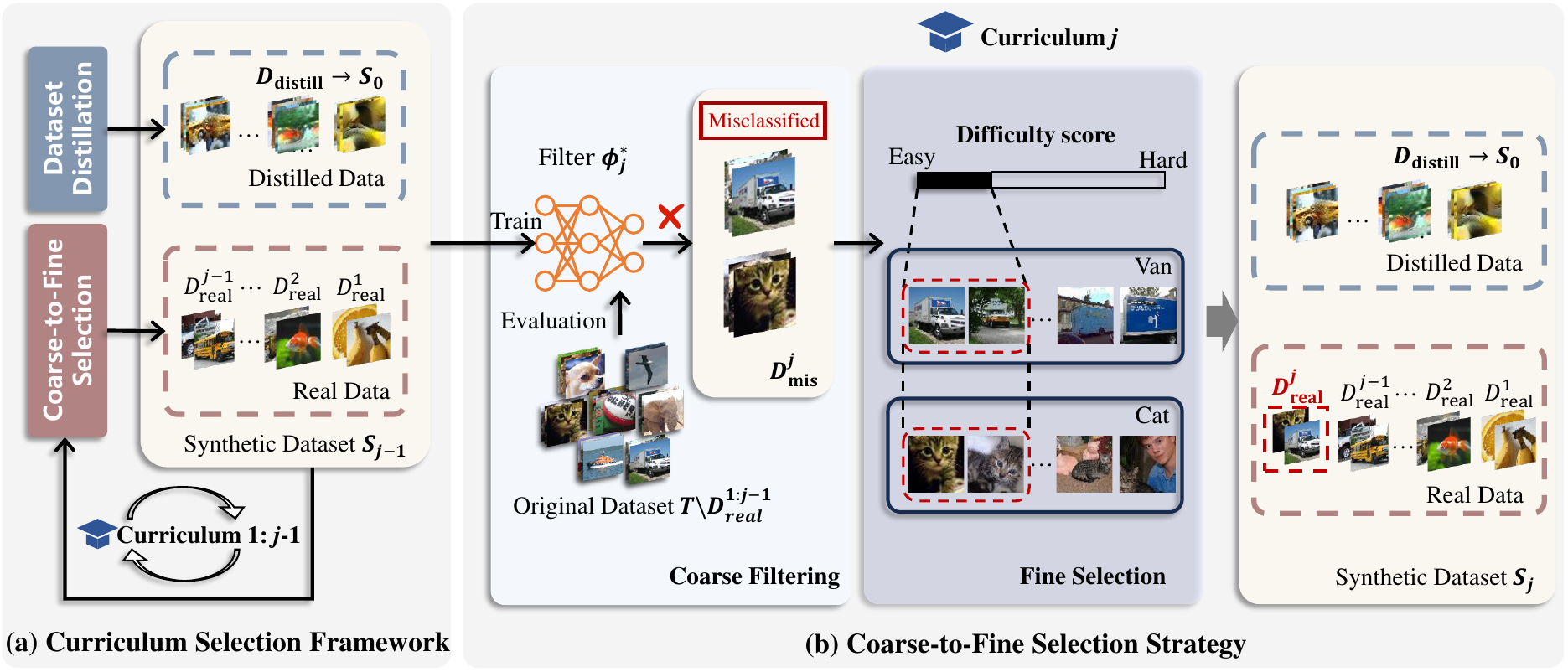}
    \caption{\textbf{Architecture of our curriculum coarse-to-fine selection method for high-IPC dataset distillation, CCFS.} CCFS adopts a combination of distilled and real data to construct the final synthetic dataset. We apply a curriculum framework and select the optimal real data for the current synthetic dataset in each curriculum. (a) \textbf{Curriculum selection framework}: CCFS begins the curriculum with the already distilled data as the initial synthetic dataset. Then continuously incorporates real data into the current synthetic dataset through the coarse-to-fine selection within each curriculum phase. (b) \textbf{Coarse-to-fine selection strategy}: In the coarse stage, CCFS trains a filter model on the current synthetic dataset and evaluates it on the original dataset excluding already selected data to filter out all correctly classified samples. In the fine stage, CCFS selects the simplest misclassified samples and incorporates them into the current synthetic dataset for the next curriculum.}
    \label{fig:architecture}
\end{figure*}
\vspace{-4mm}
\subsection{Limitations of SelMatch}\label{limiation}


The combination-based paradigm in SelMatch represents the SOTA approach for addressing the less effective problem of high-IPC dataset distillation.  However, we argue that the sliding window selection, as the core of SelMatch, may have shortcomings due to its rigid fixed and one-shot mechanism. To verify the existence of the shortcomings, we design two variants which break the mechanism. Details of the settings are as follows: 
\begin{enumerate}[label=\arabic*)] 
    \item \label{group:a} \textbf{SelMatch}: Sort the original dataset in descending order by difficulty score. Determine $\mathcal{D}_{\text{syn}}$ with a sliding window. Partition $\mathcal{D}_{\text{syn}}$ into $\mathcal{D}_{\text{real}}$ and $\mathcal{D}_{\text{distill}}$. Update $\mathcal{D}_{\text{distill}}$ by MTT's approach and keep $\mathcal{D}_{\text{real}}$ unchanged. 
    \item \label{group:b} \textbf{SelMatch w/ two-shot selection}: Change the one-shot window selection of $\mathcal{D}_{\text{real}}$ into a two-shot selection. Train a model on the initialization set $\mathcal{D}_{\text{pre-distill}}$ and evaluate it on the full training set. Select the simplest-misclassified samples and add them into $\mathcal{D}_{\text{real}}$. Repeat the process twice, then merge $\mathcal{D}_{\text{real}}$ with $\mathcal{D}_{\text{distill}}$.
    \item \label{group:c} \textbf{SelMatch w/ reverse selection}: Reverse the process in~\ref{group:b} to first conduct distillation and then make the two-shot selection. At first, conduct dataset distillation on $\mathcal{D}_{\text{pre-distill}}$ to generate $\mathcal{D}_{\text{distill}}$. Then implement the two-shot selection of $\mathcal{D}_{\text{real}}$ based on the distilled set $\mathcal{D}_{\text{distill}}$. Finally merge them to produce the synthetic dataset.
\end{enumerate}
\par\addvspace{1mm}
\noindent We conduct our analytical experiments on CIFAR-100 and evaluate the final $\mathcal{D}_{\text{syn}} = \mathcal{D}_{\textrm{real}} \cup \mathcal{D}_{\text{distill}}$ on the test dataset. The results are shown in Figure~\ref{fig:analysis_exp}.

Figure~\ref{fig:analysis_exp}(a) presents top-1 test accuracy of the 3 settings with the best hyperparameters (such as $\alpha$ and $\beta$). Across all IPC cases, setting~\ref{group:c} performs the best, followed by setting~\ref{group:b}, with setting~\ref{group:a} falling behind. Specifically,  setting~\ref{group:c} achieves an average performance improvement of 1.7\% over setting~\ref{group:a} across the four IPC settings (max: +2.8\%), while~\ref{group:b} achieves an average improvement of 0.9\% over~\ref{group:a}(max: +1.7\%). This improvement becomes more obvious as IPC increases. Figure~\ref{fig:analysis_exp}(b) presents a detailed comparison between setting~\ref{group:a} and~\ref{group:c} at various window starting point $\beta$. In all cases of $\beta$, setting~\ref{group:c} outperforms setting~\ref{group:a} and shows more stable performance fluctuations across different $\beta$.
\par\addvspace{-1mm}
In the first comparison between setting~\ref{group:a} and~\ref{group:b}, setting~\ref{group:b} obtains a better $\mathcal{D}_{\text{real}}$ through more targeted, multiple-times selections. This reflects \textbf{the rigid limitations of the fixed and one-shot sliding window initialization}. While in the second comparison between setting~\ref{group:b} and~\ref{group:c}, the only difference lies in whether the selection is made based on the initialization set $\mathcal{D}_{\text{pre-distill}}$ or the distilled set $\mathcal{D}_{\text{distill}}$. It proves that selecting based on the distilled set is better. In both~\ref{group:a} and~\ref{group:b}, although $\mathcal{D}_{\text{real}}$ is suitable for $\mathcal{D}_{\text{pre-distill}}$ at initialization, the updated $\mathcal{D}_{\text{distill}}$ by the distillation prevents the assurance of this compatibility. When we reverse the process as setting~\ref{group:c}, the selection process becomes more targeted, further enhancing the connection between $\mathcal{D}_{\text{real}}$ and $\mathcal{D}_{\text{distill}}$. \textbf{The independence between $\mathcal{D}_\textrm{real}$ and $\mathcal{D}_\textrm{distill}$ reduces the complementary effect of $\mathcal{D}_\textrm{real}$.}

These two factors collectively lead to the incompatibility issue between $\mathcal{D}_{\text{real}}$ and $\mathcal{D}_{\text{distill}}$ in the current combination paradigm, resulting in a consistent performance gap compared to full datasets. This incompatibility becomes more evident under suboptimal settings shown in Figure~\ref{fig:analysis_exp}(b). Once deviating from the optimal window position, the performance of SelMatch drops rapidly. This incompatibility issue inspired us to explore more effective strategies to select suitable real data based on the distilled data.

%% file: sec/4_method.tex
\vspace{-2mm}
\section{Method}
In this section, we introduce our Curriculum Coarse-to-Fine Selection method (CCFS) for high-IPC dataset distillation. Following the idea of combining distilled and real data to construct the synthetic dataset, CCFS aims to progressively select suitable real data based on the distilled data. We first get distilled data through dataset distillation methods. Then employ a \textbf{curriculum selection framework} for real data selection beginning with distilled data. In each curriculum, we conduct our \textbf{coarse-to-fine selection strategy} to obtain the optimal real data and integrate it with the current synthetic dataset. Figure~\ref{fig:architecture} illustrates the architecture of CCFS. 
\vspace{-7mm}
\paragraph{Curriculum Selection Framework} Our analysis experiments reveal that the fixed and one-shot selection of $\mathcal{D}_\textrm{real}$ hinders obtaining suitable real samples for the synthetic dataset. This motivated us to structure the selection process of $\mathcal{D}_\textrm{real}$ as a curriculum framework for more comprehensive coverage of essential patterns.

To make $\mathcal{D}_\textrm{real}$ a more effective complement to $\mathcal{D}_\textrm{distill}$. We choose to conduct selection after distillation is finished. We begin the curriculum with the already distilled dataset $\mathcal{D}_\textrm{distill}$ as the initial synthetic dataset $\mathcal{S}_0$ . In each curriculum phase $j$, we expect to obtain the optimal $\mathcal{D}_{\textrm{real}}^{j}$ from the original dataset for the current synthetic dataset $\mathcal{S}_{j-1}$:
\begin{equation}\label{eq4}
    \begin{aligned}
        \mathcal{D}_{\textrm{real}}^{j} = \operatorname{Select}(\mathcal{S}_{j-1}, \mathcal{T} \backslash \mathcal{D}_{\textrm{real}}^{1:j-1}) \\
        \text{s.t. } \mathcal{S}_{0} = \mathcal{D}_{\textrm{distill}} \textrm{ and } j \in \{1, 2, \ldots, J\},
    \end{aligned}
\end{equation}
where $J$ is the number of curricula and $\operatorname{Select}$ denotes the selection strategy used in each curriculum. Note that previously selected samples are excluded in each curriculum.

Then we incorporate $\mathcal{D}_{\textrm{real}}^{j}$ into $\mathcal{S}_{j-1}$ as $\mathcal{S}_{j}$ for the next curriculum phase. After the last curriculum, we obtain the final synthetic dataset $\mathcal{S}$ with the target size: 
\par\addvspace{-2mm}
\begin{equation}\label{eq5}
   \mathcal{S}_{j} = \mathcal{S}_{j-1} \cup \mathcal{D}_{\textrm{real}}^{j},
\end{equation}
\begin{equation}\label{eq6}
    \mathcal{S} = \mathcal{S}_{J}.
\end{equation}
\par\addvspace{-1mm}
We expect to gradually incorporate suitable features through this curriculum selection framework. Next, we introduce our strategy of selecting the optimal $\mathcal{D}_\textrm{real}$ in each curriculum phase and explain how this strategy functions within the curriculum framework.
\paragraph{Coarse-to-Fine Selection Strategy}  The weak connection between $\mathcal{D}_\textrm{real}$ and $\mathcal{D}_\textrm{distill}$ in the incompatibility issue calls for a more targeted selection strategy. We design a two-step strategy, coarse-to-fine selection, to get the optimal $\mathcal{D}_\textrm{real}$ in each curriculum phase. Given a synthetic dataset $\mathcal{S}$ , we first use a filter $\phi$ trained on $\mathcal{S}$ to evaluate on the original training dataset $\mathcal{T}$ and obtain the misclassified samples $\mathcal{D}_\textrm{mis}$:
\begin{equation}\label{eq7}
    \mathcal{D}_{\textrm{mis}} = \left\{ (x_i, y_i) \in \mathcal{T} \mid \phi_{\bm{\theta}_{\mathcal{S}}}^*(x_i) \neq y_i \right\},
\end{equation}

\noindent where $\bm{\theta}_{\mathcal{S}}$ denotes the parameters of  $\phi$ optimized on $\mathcal{S}$. 

This evaluation can coarsely reflect the limitations of current $\mathcal{S}$. These limitations are primarily concentrated in the misclassified samples $\mathcal{D}_\textrm{mis}$ from the original training set. Considering $\mathcal{D}_\textrm{mis}$ may include samples with beneficial, harmful, or negligible influence on training~\cite{koh2017understanding,pruthi2020estimating}, we conduct a finer selection next. We arrange the samples of  $\mathcal{D}_\textrm{mis}$ in ascending order based on pre-computed difficulty scores~\cite{toneva2018forgetting,jiang2020cscore} and then select the easiest samples up to the target size as the optimal complement to current $\mathcal{S}$:
\begin{equation}\label{eq8}
    \mathcal{D}_{\textrm{real}} = \bigcup_{c \in \mathcal{C}} \left\{ (x_i, y_i) \in \mathcal{D}_{\textrm{mis}}^{(c)}\,\middle|\, \operatorname{rank}_{\mathcal{D}_{\textrm{mis}}^{(c)}}(x_i) \leq k \right\},
\end{equation}
\noindent where $\mathcal{C}$ represents the total number of classes, $\mathcal{D}_{\textrm{mis}}^{(c)}$ includes samples of class $c$ in $\mathcal{D}_\textrm{mis}$, $\operatorname{rank}$ denotes the ascending order arrangement of sample difficulty, and $k$ is the target complement amount for each class. To keep balance, we select an equal number of complement samples per class. 

We figure that simple features that haven't been learned are essential for $\mathcal{S}$. The nature of dataset distillation often leads to synthesizing mainly easy and representative features from the original dataset~\cite{guo2024lossless,lee2024selmatch,ma2024cudd}, which provides the filter with fundamental classification capabilities as evidenced by correctly classified samples. In the first step, we coarsely filter out correctly classified samples to avoid reintroducing these already learned simple features.

Now we have $\mathcal{D}_\textrm{mis}$ that reflects the limitations in $\mathcal{S}$. Among these limitations, simpler features provide greater benefit to model training compared to more difficult and complex features, as they are easier to learn.  Pre-calculated difficulty scores effectively measure the relative difficulty of sample features from a global perspective, guiding our fine selection in the next step. By selecting the simplest ones from misclassified samples, we obtain the optimal $\mathcal{D}_{\textrm{real}}$ while avoiding the introduction of overly complex features that could hinder the performance of $\mathcal{S}$. Section~\ref{ablation:strategy} further demonstrates the effectiveness of our selection strategy.

Algorithm~\ref{alg:framework} describes the entire process of the CCFS algorithm. We employ this coarse-to-fine selection in our curriculum framework. Although we repeatedly select the simplest-misclassified samples across curriculum phases, the continuous enrichment of the synthetic dataset leads to an improvement in the filter's capacity, which in turn raises the lower bound of the difficulty for misclassified samples. Meanwhile, the strategy of selecting the simplest samples maintains a manageable difficulty progression between curriculum phases. We provide more details on the progressive nature of CCFS in Further Analysis~\ref{further}.

\begin{algorithm}[ht]
    \caption{\algopt: A curriculum coarse-to-fine selection framework for high-IPC dataset distillation}\label{alg:framework}
    \begin{algorithmic}
        \Statex \textbf{Input:} Original full dataset $\cT$, number of classes $C$, target images per class \IPC, distillation portion $\alpha \in [0, 1]$, dataset distillation algorithm $\cA$, number of curricula $J$, pre-calculated difficulty score $\bm{score}$. \\
        \State $\cD_\textrm{distill} = \cA(\cT)$, s.t. $|\cD_\textrm{distill}| = \lceil \alpha \times \IPC \times C \rceil$ 
        \State $\cS_0 \gets \cD_\textrm{distill}$ 
        
        \For{$j = 1$ {\bfseries to} $J$}
        \State $k_j = \lfloor \frac{ \IPC \times (1 - \alpha)}{J} \rfloor$ 
        \State Train the filter model on $\cS_{j-1}$ to get $\phi_j^*$
        \State $\triangleright$ {\textcolor{blue}{Coarse Filtering}}
        \State $\cD_\textrm{mis}^j = \left\{ (x_i, y_i) \in \mathcal{T} \backslash \cD_\textrm{real}^{1:j-1} \mid \phi_j^*(x_i) \neq y_i \right\}$ 
        \State $\triangleright$ {\textcolor{blue}{Fine Selection}}
        \State $\operatorname{rank}_{\cD_{\textrm{mis}}^{j, c}} \gets \operatorname{sort}_{\text{asc}}(\cD_{\textrm{mis}}^{j,c}, \bm{score})$ 
        \State $\mathcal{D}_{\textrm{real}}^j = \displaystyle\bigcup_{c \in \mathcal{C}} \left\{ (x_i, y_i) \in \mathcal{D}_{\textrm{mis}}^{j, c}\,\middle|\, \operatorname{rank}_{\cD_\textrm{mis}^{j, c}}(x_i) \leq k_j \right\}$ 
        \State $\cS_j \gets \cS_{j-1} \cup \mathcal{D}_{\textrm{real}}^j$ 
        \EndFor
        \Statex \textbf{Output:} The final synthetic dataset $\cS \gets \cS_J$
    \end{algorithmic}
\end{algorithm}

%% file: sec/5_experiment.tex
\begin{table*}[t]
\captionsetup{position=bottom}
\setlength{\abovecaptionskip}{1mm}
\setlength{\belowcaptionskip}{0mm}
\centering
\caption{\textbf{Performance of CCFS compared to the SOTA dataset distillation and coreset selection baselines.} We report the results of all listed methods with the identical validation model ResNet-18. CCFS achieves state-of-the-art performance across high-IPC settings ranging from 5\% to 30\% compression ratio. Additionally, the selection-only version of our method, self-evolved selection, beats other coreset selection baselines and exhibits comparable performance to SOTA dataset distillation methods. $^*$ denotes results obtained using the official code due to the incomplete results shown in original papers. \textit{IPC}: images per class, \textit{Ratio}: the compression ratio of the synthetic dataset compared  to the original dataset.}
\label{tab:main_result}
\begin{adjustbox}{max width=\linewidth}
\begin{threeparttable}
\begin{tabular}{c|cccc|cccc|cc}
\toprule
Dataset  & \multicolumn{4}{c|}{CIFAR-10} & \multicolumn{4}{c|}{CIFAR-100} & \multicolumn{2}{c}{Tiny-ImageNet} \\
IPC   & 250 & 500 & 1000 & 1500 & 25 & 50 & 100 & 150 & 50 & 100 \\
Ratio & $5\%$  & $10\%$ & $20\%$ &$30\%$ & $5\%$  & $10\%$ & $20\%$ & $30\%$ & $10\%$ & $20\%$ \\
\midrule
Random & 73.4$\pm$1.5 & 79.3$\pm$0.3 & 85.6$\pm$0.4 & 88.3$\pm$0.2 & 35.8$\pm$0.6 & 40.7$\pm$1.0 & 53.2$\pm$0.9 & 60.3$\pm$1.3 & 30.1$\pm$0.6 & 40.1$\pm$0.4 \\
Forgetting~\cite{toneva2018forgetting} & 30.7$\pm$0.3 & 41.5$\pm$0.7 & 68.4$\pm$1.6 & 83.5$\pm$1.8 & 9.5$\pm$0.3 & 13.2$\pm$0.6 & 27.0$\pm$1.1 & 42.3$\pm$1.0 & 5.7$\pm$0.1 & 12.4$\pm$0.2 \\
Glister~\cite{killamsetty2021glister} & 46.6$\pm$1.3 & 56.6$\pm$0.5 & 79.0$\pm$0.7 & 85.0$\pm$0.9 & 21.7$\pm$0.8 & 26.7$\pm$1.3 & 39.9$\pm$1.4 & 52.1$\pm$1.3 & 22.6$\pm$0.5 & 34.0$\pm$0.3 \\
Oracle window~\cite{lee2024selmatch} & 79.3$\pm$0.7 & 85.2$\pm$0.1 & 89.9$\pm$0.5 & 90.6$\pm$0.3 & 43.2$\pm$1.8 & 50.0$\pm$0.8 & 59.2$\pm$0.8 & 64.7$\pm$0.5 & 42.5$\pm$0.3 & 49.2$\pm$0.3 \\ 
Self-evolved selection & \underline{81.6$\pm$0.5} & \underline{86.4$\pm$0.3} & \underline{90.3$\pm$0.5} & \underline{91.6$\pm$0.4} & \underline{45.6$\pm$0.5} & \underline{50.7$\pm$0.7} & \underline{62.6$\pm$0.8} & \underline{66.5$\pm$0.2} & \underline{43.9$\pm$0.6} & \underline{50.2$\pm$0.4} \\ 
\midrule
DSA~\cite{zhao2021DSA} & 74.7$\pm$1.5 & 78.7$\pm$0.7 & 84.8$\pm$0.5 & - & 38.4$\pm$0.4 & 43.6$\pm$0.7 & - & - & 27.8$\pm$1.4 & - \\
DM~\cite{zhao2023dm} & 75.3$\pm$1.4 & 79.1$\pm$0.6 & 85.6$\pm$0.5 & - & 37.5$\pm$0.6 & 42.6$\pm$0.5 & - & - & 31.0$\pm$0.6 & - \\
MTT~\cite{cazenavette2022mtt} & 80.7$\pm$0.4 & 82.2$\pm$0.4 & 86.1$\pm$0.3 & 88.6$\pm$0.2 & 49.9$\pm$0.7 & 51.3$\pm$0.4 & 58.7$\pm$0.6 & 63.1$\pm$0.3 & 40.3$\pm$0.3 & 44.2$\pm$0.5 \\
$\text{SRe}^2\text{L}^*$~\cite{yin2023sre2l}& 77.5$\pm$0.7 & 85.1$\pm$0.2 & 86.8$\pm$0.3 & 87.8$\pm$0.4 & 49.7$\pm$0.4 & 51.4$\pm$0.4 & 58.8$\pm$0.2 & 61.9$\pm$0.3 & 41.1$\pm$0.4 & 49.7$\pm$0.3  \\ 
DATM~\cite{guo2024lossless} & - & 84.8$\pm$0.3 & 87.6$\pm$0.3 & - & - & 51.0$\pm$0.5 & 61.5$\pm$0.3 & - & 42.2$\pm$0.2 & - \\
SelMatch~\cite{lee2024selmatch} & 82.8$\pm$0.2 & 85.9$\pm$0.2 & 90.4$\pm$0.2 & 91.3$\pm$0.2 & 50.9$\pm$0.3 & 54.5$\pm$0.6 & 62.4$\pm$0.5 & 67.4$\pm$0.2 & 44.7$\pm$0.2 & 50.4$\pm$0.2 \\
CDA$^*$ ~\cite{yin2023cda}& 78.0$\pm$0.4 & 84.4$\pm$0.4 & 86.4$\pm$0.2 & 87.5$\pm$0.4 & 50.6$\pm$0.3 & 59.7$\pm$0.2 & 61.1$\pm$0.1 & 63.4$\pm$0.2 & 45.6$\pm$0.2 & 52.4$\pm$0.1 \\
CUDD~\cite{ma2024cudd} & - & - & - & - & 63.5$\pm$0.3 & 65.7$\pm$0.2 & - & - & 55.6$\pm$0.2 & 56.8$\pm$0.2 \\
CCFS (Ours)  & \textbf{87.9}$\pm$\textbf{0.4} & \textbf{92.5}$\pm$\textbf{0.2} & \textbf{93.2}$\pm$\textbf{0.1} & \textbf{93.8}$\pm$\textbf{0.1} & \textbf{65.3}$\pm$\textbf{0.2} & \textbf{71.5}$\pm$\textbf{0.3} & \textbf{73.0}$\pm$\textbf{0.2} & \textbf{74.8}$\pm$\textbf{0.2} & \textbf{55.8}$\pm$\textbf{0.3} & \textbf{60.2}$\pm$\textbf{0.2} \\ 
\midrule
Full Dataset & \multicolumn{4}{c|}{95.5$\pm$0.2}   & \multicolumn{4}{c|}{78.8$\pm$0.3} & \multicolumn{2}{c}{60.5$\pm$0.2}  
\\ \bottomrule
\end{tabular}
\end{threeparttable}
\end{adjustbox}
\end{table*}

\section{Experimental Results}
\label{sec:experiments}

\subsection{Experiment Setup}
We evaluate the performance of our method CCFS on various datasets including CIFAR-10, CIFAR-100, and Tiny-ImageNet. We compare our method with SOTA dataset distillation and coreset selection methods. For coreset selection baselines, we include Glister~\cite{killamsetty2021glister}, Forgetting~\cite{toneva2018forgetting}, and the oracle-window selection proposed in SelMatch. For comparison, we also report a selection-only version of CCFS, referred to as self-evolved selection. For dataset distillation baselines, we incorporate DSA~\cite{zhao2021DSA}, DM~\cite{zhao2023dm}, MTT~\cite{cazenavette2022mtt}, DATM~\cite{guo2024lossless}, SelMatch~\cite{lee2024selmatch}, $\text{SRe}^2\text{L}$~\cite{yin2023sre2l}, CDA~\cite{yin2023cda} and CUDD~\cite{ma2024cudd}. We also report the full dataset training performance with 200 training epochs.
\vspace{-2mm}
\paragraph{Datasets Details.}
\begin{itemize}
    \item CIFAR-10~\cite{krizhevsky2009cifar}: 10 classes with 5000 low-resolution (32 $\times$ 32) training images per class, 10000 images for testing.
    \item CIFAR-100~\cite{krizhevsky2009cifar}: 100 classes with 500 low-resolution (32 $\times$ 32) training images per class, 10000 images for testing.
    \item Tiny-ImageNet~\cite{le2015tiny}: 200 classes with 500 high-resolution (64 $\times$ 64) training images per class, 10000 images for validation.
\vspace{-2mm}
\end{itemize}
\paragraph{Evaluation Networks.} We choose ResNet-18~\cite{he2016resnet} as the uniform evaluation network for the main comparison. For cross-architecture generalization, we use ResNet-50/101, DenseNet-121~\cite{huang2017densnet}, and RegNet-Y-8GF~\cite{radosavovic2020regnet} as the evaluation backbones.

\paragraph{Implement Details of CCFS.} We begin with $\mathcal{D}_\textrm{distill}$ synthesized by CDA~\cite{yin2023cda} method, which belongs to $\text{SRe}^2\text{L}$~\cite{yin2023sre2l} series and implements a progressive difficulty data augmentation on the synthetic images during distillation. In the next curriculum selection, we set the default number of curriculum phases to 3 and evenly distribute the samples to be selected among them. In each curriculum, we train ResNet-18 from scratch on the current synthetic dataset as the filter, using equal training epochs as those in the final evaluation. We use pre-calculated Forgetting~\cite{toneva2018forgetting} scores and apply our selection strategy on the training set, excluding previously selected samples. We report the results of the optimal distillation portion $\alpha$ at each IPC setting. Our method has excellent scalability and can be adapted to various dataset distillation methods. We present the results of combining CCFS with MTT~\cite{cazenavette2022mtt} dataset distillation in the Appendix.
\subsection{Main Results}
We compare our method with the state-of-the-art dataset distillation and coreset selection methods on CIFAR-10, CIFAR-100, and Tiny-ImageNet under high-IPC settings ranging from 5\% to 30\% compression ratios. As shown in Table~\ref{tab:main_result}, previous distillation methods gradually lose effectiveness as IPC increases, even falling behind random selection. By progressively introducing suitable real samples into the synthetic dataset, CCFS establishes new state-of-the-art performance in high-IPC settings. Notably, our method achieves a performance gain of \{6.6\%, 5.8\%\} on \{CIFAR-10, CIFAR-100\} with compression ratio of 10\%. For Tiny-ImageNet with IPC=100 (20\% compression ratio), we achieve 60.2\% top-1 test accuracy, representing a 3.4\% improvement over the current state-of-the-art method. This performance comes remarkably close to the 60.5\% test accuracy of full dataset training.

Additionally, we report a selection-only version of CCFS, referred to as self-evolved selection. We select real samples of appropriate difficulty with a sliding window as the initial coreset and expand it following the CCFS strategy. Self-evolved selection significantly outperforms other coreset selection methods in high IPC and also exhibits comparable performance to advanced dataset distillation approaches. It indicates that progressively selecting suitable real images is crucial for producing a coreset with the largest coverage of essential patterns in the original dataset.

\begin{table}[t]
\setlength{\abovecaptionskip}{1mm}
\setlength{\belowcaptionskip}{-4mm}
\centering\scriptsize
\caption{Cross-architecture experiment results on Tiny-ImageNet with IPC=100.}
\label{tab:cross}
\begin{tabular}{@{}cccccc@{}}
\toprule
\multirow{2}{*}{Method}& \multicolumn{5}{c}{Validation Model} \\ \cmidrule(l){2-6}
              & R18   & R50   & R101  & DenseNet-121 & RegNet-Y-8GF  \\ \midrule
$\text{SRe}^2\text{L}$   & 48.00 & 51.02 & 51.92 & 50.66 & 54.78 \\
CDA                 & 51.12 & 54.00 & 55.04 & 52.47 & 57.13 \\
CCFS (Ours)         & \textbf{60.20} & \textbf{60.67} & \textbf{61.17} & \textbf{60.52} & \textbf{62.94} \\ \bottomrule
\end{tabular}
\end{table}

\begin{table}[t]
\setlength{\abovecaptionskip}{1mm}
\setlength{\belowcaptionskip}{-2mm}
\centering\scriptsize
\caption{Ablation study on the select strategy on CIFAR-100 with IPC=50.}
\label{tab:ablation_select_method}
\begin{tabular}{@{} cccc @{}}
\toprule
\multirow{2}{*}{Coarse Stage} & \multicolumn{3}{c}{Fine Stage} \\ \cmidrule(l){2-4}
& Simple & Hard & Random \\ 
\midrule
Classified & 66.8 & 63.5 & 66.8 \\
Misclassified & \textbf{71.5} & 65.0 & 70.1 \\
\bottomrule
\end{tabular}
\vspace{-3mm}
\end{table}

\subsection{Cross-architecture Generalization}
To further evaluate CCFS’s effectiveness, we conduct cross-architecture generalization experiments using additional validation models beyond the ResNet-18 in the main table, including ResNet-50/101, DenseNet-121, and RegNet-Y-8GF. We set ResNet-18 as the filter model for curriculum selection and generate the final synthetic dataset, which is then used to train other validation models from scratch. As shown in Table~\ref{tab:cross}, our method demonstrates robust generalization performance.

\begin{figure*}[ht]
    \centering
        \setlength{\abovecaptionskip}{-2mm}
        \setlength{\belowcaptionskip}{0mm}
        \includegraphics[width=1\linewidth]{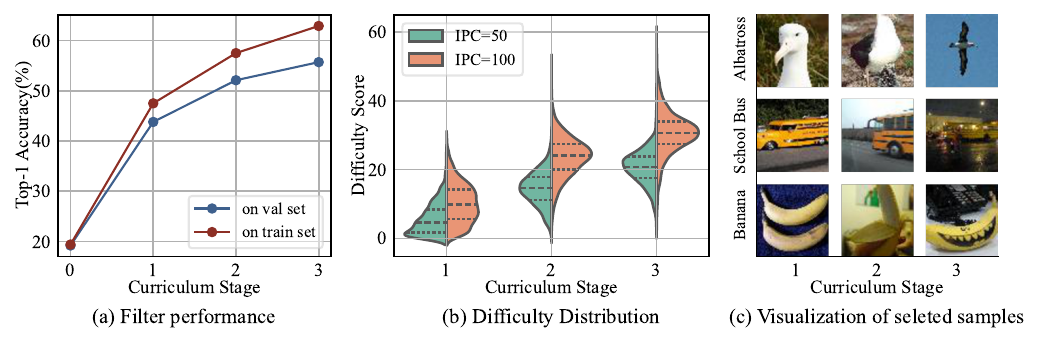}
    \caption{\textbf{Further analysis on the curriculum framework.} (a) Performance of the filter model trained on the synthetic dataset in each curriculum phase with IPC=50: The filter’s classification accuracy steadily improves on both the original training set and the validation set. (b) The difficulty distribution of real samples selected in each curriculum phase: As the curriculum progresses, both the average difficulty as well as the upper and lower difficulty bounds of selected samples increase significantly. Moreover, higher IPC tend to include more difficult samples than lower IPC within the same curriculum phase. CCFS effectively guides the synthetic dataset to incorporate more challenging samples. (c) Visualization of the samples selected in each curriculum phase. We present images of median difficulty across several categories in Tiny-ImageNet: Albatross, School Bus and Banana. The visualization effectively illustrates the gradual increase in difficulty (diverse poses, complex backgrounds, other distractions...) facilitated by CCFS. }
    \label{fig:analysis on curriculum}
    \vspace{-2mm}
\end{figure*}

\subsection{Ablation Study}

\paragraph{The selection strategy.}\label{ablation:strategy} In our coarse-to-fine selection strategy, we choose the misclassified subset in the coarse stage and select the simplest ones in the fine stage next. This combination has demonstrated excellent performance. Here, we explore other combinations. Specifically, for each curriculum selection, we select the simplest/hardest or just randomly select samples from correctly-classified/misclassified subset by current trained filter model. We conducted experiments on CIFAR-100 with IPC=50. Among all results shown in Table~\ref{tab:ablation_select_method}, the simplest-misclassified selection strategy outperforms all other combinations, further demonstrating its effectiveness.

\paragraph{The difficulty scores.} Our method requires difficulty scores to measure the complexity of samples. We explored the impact of using different difficulty scores. We include pre-calculated C-score~\cite{jiang2020cscore} and Forgetting scores~\cite{toneva2018forgetting}. Additionally, we measure sample difficulty using the predicted values (logits) by the current trained filter model for the actual class of each training sample, with smaller values indicating greater difficulty. We conducted experiments on CIFAR-10, CIFAR-100, and Tiny-ImageNet with a 10\% compression ratio using these three scores. Results in Table~\ref{tab:ablation_score} indicate that the Forgetting score outperformed the others across all three datasets, while the logits approach leads to performance degradation due to its coarse reflection of sample difficulty. The Forgetting score is leveraged for the main results (Table~\ref{tab:main_result}).\vspace{-2mm}
\paragraph{The number of curricula.}\label{ablation:curriculum_num} Table~\ref{tab:curriculum_number} illustrates the impact of varying the number of curricula. It indicates that moderate increases improve performance, further demonstrating the effectiveness of the curriculum selection framework. However, continuing to increase number of curricula results in only marginal performance gains. Balancing performance and efficiency, we employ 3 curriculum phases  in our main results (Table~\ref{tab:main_result}).

\begin{table}[t]
\setlength{\abovecaptionskip}{1mm}
\setlength{\belowcaptionskip}{-4mm}
\centering\scriptsize
\caption{Ablation study on the difficulty score used in selection strategy with 10\% compression ratio on CIFAR-10, CIFAR-100 and Tiny-ImageNet.}
\label{tab:ablation_score}
\begin{tabular}{@{}c|ccc@{}}
\toprule
Score & CIFAR-10 & CIFAR-100 & Tiny-ImageNet \\
\midrule
Logits & 91.8 & 68.7 & 52.5\\
C-score & 92.2 & 71.0 & -\\
Forgetting & \textbf{92.5} & \textbf{71.5} & \textbf{55.8}\\
\bottomrule
\end{tabular}
\end{table}

\begin{table}[t]
\setlength{\abovecaptionskip}{1mm}
\setlength{\belowcaptionskip}{-1mm}
\centering\scriptsize
\caption{Ablation study on the number of curricula with 10\% compression ratio on CIFAR-10, CIFAR-100 and Tiny-ImageNet.}
\label{tab:curriculum_number}
\begin{tabular}{@{}c|ccc@{}}
\toprule
Number of curricula & CIFAR-10 & CIFAR-100 & Tiny-ImageNet \\
\midrule
1 & 91.6 & 67.9 & 54.4\\
2 & 91.8 & 70.4 & 55.3\\
3 & 92.5 & 71.5 & 55.8\\
4 & 92.4 & 71.6 & 55.7\\
\bottomrule
\end{tabular}
\vspace{-3mm}
\end{table}

\subsection{Further Analysis}\label{further}
The curriculum framework in CCFS aims to progressively incorporate suitable samples into the synthetic dataset, thereby enhancing its performance incrementally. We expect to observe a continuous improvement in the classification capability of the filter model, along with a gradual increase in the overall difficulty of the selected samples as the curriculum progresses. To verify that CCFS achieves these intended effects, we conduct extensive experiments on Tiny-ImageNet with 3 curriculum phases and record the state in each curriculum phase. Figure~\ref{fig:analysis on curriculum} illustrates the effectiveness of the curriculum framework in CCFS from both the filter performance and sample difficulty.

In Figure~\ref{fig:analysis on curriculum}(a), we present the performance of the filter model trained on the synthetic dataset in each curriculum phase with IPC=50. As the curriculum progresses, the filter’s classification accuracy steadily improves on both the original training set and the validation set. This indirectly reflects the growing representational capacity of the synthetic dataset. 

Figure~\ref{fig:analysis on curriculum}(b) illustrates the difficulty distribution of real samples selected in each curriculum phase. We can observe that the selected samples' average difficulty increases significantly across 3 curriculum phases as intended. Moreover, as the curriculum progresses, both the upper and lower bounds of the difficulty in selected samples are rising. This trend indicates that CCFS effectively enhances the overall difficulty of the synthetic dataset instead of struggling within a similar range. Since we continuously select the simplest-misclassified samples in each curriculum phase, the increase in lower bounds also indirectly reflects the rising threshold for the filter model's errors, indicating a more concrete  improvement in its filtering capability, rather than just a simple performance boost. Consequently, CCFS effectively guides the synthetic dataset to incorporate more challenging and training-valuable samples. Additionally, within the same curriculum phase, synthetic datasets with higher IPC tend to include more difficult samples. This also aligns with our expectations: larger synthetic datasets are supposed to encapsulate more complex information. As IPC increases, CCFS successfully introduces harder and rarer features into the synthetic dataset.

In Figure~\ref{fig:analysis on curriculum}(c), we visualize the samples selected in different curriculum phases, showcasing samples of median difficulty across several categories. In the early stages, CCFS tends to select classic samples that capture the general features of the category. These images have simple backgrounds and fully visible objects. As the curriculum progresses, more challenging samples are incorporated into the synthetic dataset, featuring diverse poses (e.g., bird in flight, peeled banana), partial views (e.g., the lower half of bird, the front of school bus), complex backgrounds, and other distractions. This visualization effectively illustrates the gradual increase in difficulty facilitated by CCFS.

%% file: sec/6_conclusion.tex
\section{Conclusion}
In this paper, we reveal the incompatibility issue between distilled and real data in the current combination-based dataset distillation method through a series of analysis experiments. We propose CCFS, a novel combination-based framework for high-IPC dataset distillation. We apply a curriculum selection framework for real data and begin the curriculum with distilled data. This ensures suitable features are progressively introduced into the synthetic dataset across curriculum phases. In each curriculum phase, we employ our coarse-to-fine selection strategy to obtain the optimal real data for the current synthetic dataset. This effectively enhances the connection between distilled and real data. CCFS significantly narrows the performance gap between synthetic datasets and full datasets under high-IPC conditions. We achieve state-of-the-art performance in various high-IPC settings on CIFAR-10, CIFAR-100, and Tiny-ImageNet. Further analyses demonstrate the effectiveness of our selection strategy and the expected progressive effect of the curriculum framework. CCFS also exhibits robust cross-architecture generalization and excellent scalability  to other distillation approaches.

%% file: sec/7_acknowledgement.tex
\section*{Acknowledgement}
This study is supported by the National Natural Science Foundation of China (Grant No. 62306084, U23B2051, 62476071, and U24A20328), and Shenzhen Science and Technology Program (Grant No. GXWD20231128102243003, KJZD20230923115113026, and ZDSYS20230626091203008), and China Postdoctoral Science Foundation (Grant No. 2024M764192).

%% file: sec/X_appendix.tex

\appendix

\section{Implement Details}
\label{app:details}
\vspace{-2mm}
In this section, we introduce more details about the implementation of CCFS. In the main results, we choose CDA as the base distillation method. We utilize its official code to synthesize $\cD_\textrm{distill}$ for CIFAR-10/100 and Tiny-ImageNet. We also leverage the pre-generated soft label approach for the final synthetic data as CDA. Here, we don't elaborate on details of the dataset distillation. We provide implementation details of the subsequent curriculum selection and the final evaluation below.
\vspace{-1mm}
\subsection{CIFAR-10/100}
\par\addvspace{-2mm}
\paragraph{Hyper-parameter Setting.} In curriculum selection, we set the default number of curriculum phases to 3 and evenly distribute the samples to be selected among them. In each curriculum, we train a modified ResNet-18 model from scratch on the current synthetic dataset as the filter, using equal training epochs as those in the final evaluation. We use pre-calculated Forgetting scores and apply our coarse-to-fine selection strategy to the training set, excluding previously selected samples. For evaluation, we train the identical ResNet-18 on the final synthetic dataset and follow the same training settings as the filter. The hyperparameter settings are shown in Table~\ref{tab:details_cifar}.

\begin{table}[h]
    \setlength{\abovecaptionskip}{1mm}
    \setlength{\belowcaptionskip}{-3mm}
    \centering
    \caption{Hyperparameter settings on CIFAR-10/100.}
    \label{tab:details_cifar}
    \begin{tabular}{@{}l|l@{}}
    \toprule
    config                 & value            \\ \midrule
    difficulty score       & Forgetting \\
    number of curricula    & 3          \\
    optimizer              & SGD         \\
    base learning rate     & 0.1             \\
    momentum               & 0.9  \\
    weight decay           & 5e-4             \\
    learning rate schedule & cosine decay \\
    augmentation           & RandomResizedCrop   \\
    \bottomrule
    \end{tabular}
\vspace{-2mm}
\end{table}

For the hyperparameter---training epochs, we set the same training epochs for both the filter and the final evaluation model. The number of training epochs varies based on the target IPC. We assign more training epochs to smaller IPC settings, following the settings in other dataset distillation methods. Table~\ref{tab:epochs_cifar} shows the specific settings of the training epochs.

For the hyperparameter---batch size, We configure it based on the current size of the synthetic dataset considering its progressive growth across curriculum phases. As the size of the synthetic dataset grows, we appropriately increase the batch size in the filter training. For evaluation, we similarly set the evaluation model’s training batch size based on the size of the final synthetic dataset. Table~\ref{tab:bs_cifar} presents the detailed settings for the batch size.

\begin{table}[h]
    \setlength{\abovecaptionskip}{1mm}
    \centering
    \caption{Training epochs configuration on CIFAR-10/100.}
    \label{tab:epochs_cifar}
    \begin{tabular}{@{}l|cccc@{}}
    \toprule
    Compression Ratio & 5\% & 10\% & 20\% & 30\%  \\ \midrule
    Training Epochs   & 500 & 500 & 250 & 200   \\
    \bottomrule
    \end{tabular}
\end{table}

\begin{table}[h]
    \centering
    \caption{Batch size configuration for both the filter and the evaluation training according to the size of the current on CIFAR-10/100.}
    \label{tab:bs_cifar}
    \begin{tabular}{@{}l|ccc@{}}
    \toprule
    Compression Ratio & $\leq5\%$ & $5\%-20\%$ & $>20\%$ \\ \midrule
    Batch Size   & 32 & 64 & 128   \\
    \bottomrule
    \end{tabular}
    \vspace{-2mm}
\end{table}

\subsection{Tiny-ImageNet}
\paragraph{Hyper-parameter Setting.} In curriculum selection, we set the default number of curriculum phases to 3 and evenly distribute the samples to be selected among them. In each curriculum, we train a modified ResNet-18 model from scratch on the current synthetic dataset as the filter, using equal training epochs as those in the final evaluation. We use pre-calculated Forgetting scores and apply our coarse-to-fine selection strategy to the training set, excluding previously selected samples. For evaluation, we train the identical ResNet-18 on the final synthetic dataset and follow the same training settings as the filter. We uniformly set the training epochs to 100 and the batch size to 64 for both the filter training across curriculum phases and the final evaluation. The hyperparameter settings are shown in Table~\ref{tab:details_tiny}.

\begin{table}[h]
    \centering
    \caption{Parameter setting on CIFAR-10/100.}
    \label{tab:details_tiny}
    \begin{tabular}{@{}l|l@{}}
    \toprule
    config                 & value            \\ \midrule
    difficulty score       & Forgetting \\
    number of curricula    & 3          \\
    optimizer              & SGD         \\
    base learning rate     & 0.2             \\
    momentum               & 0.9  \\
    weight decay           & 1e-4             \\
    learning rate schedule & cosine decay \\
    augmentation           & RandomResizedCrop   \\
    training epochs        & 100   \\
    batch size             & 64 \\
    \bottomrule
    \end{tabular}
    \vspace{-1mm}
\end{table}

\begin{figure*}[t]
    \centering
        \setlength{\abovecaptionskip}{-2mm}
        \setlength{\belowcaptionskip}{-2mm}
        \includegraphics[width=1\linewidth]{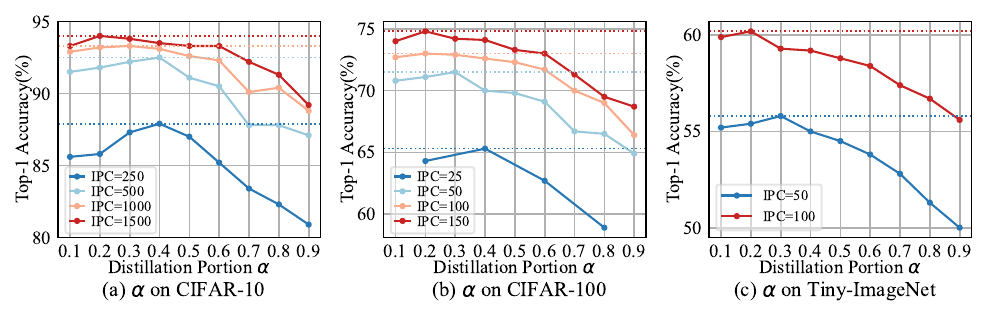}
    \caption{\textbf{Impact of different distillation portion $\alpha$ on CIFAR-10/100 and Tiny-ImageNet.}  We recommend a small distillation portion $\alpha$ in high-IPC settings.}
    \label{fig:alpha}
\end{figure*}

\begin{table*}[h]
\captionsetup{position=bottom}
\setlength{\abovecaptionskip}{2mm}
\setlength{\belowcaptionskip}{-3mm}
\centering
\begin{threeparttable}
\begin{tabular}{@{}c|cccc|cccc@{}}
\toprule
Dataset  & \multicolumn{4}{c|}{CIFAR-10} & \multicolumn{4}{c}{CIFAR-100} \\
IPC   & 250 & 500 & 1000 & 1500 & 25 & 50 & 100 & 150 \\
Ratio & $5\%$  & $10\%$ & $20\%$ &$30\%$ & $5\%$  & $10\%$ & $20\%$ & $30\%$ \\
\midrule
SelMatch & 82.8 & 85.9 & 90.4 & 91.3 & 50.9 & 54.5 & 62.4 & 67.4 \\
CCFS w/ MTT  & \textbf{83.2} & \textbf{86.3} & \textbf{91.0} & \textbf{92.1} & \textbf{51.6} & \textbf{56.0} & \textbf{65.2} & \textbf{69.2} \\ 
\bottomrule
\end{tabular}
\end{threeparttable}
\caption{\textbf{Results of CCFS with MTT as the base dataset distillation method.} CCFS with MTT still outperforms SelMatch across all high-IPC settings, showcasing its excellent scalability.}
\label{tab:mtt}
\end{table*}

\section{Distillation Portion}
The portion $\alpha$ of $\cD_\textrm{distill}$ in the final synthetic dataset is another key hyperparameter. In the main table, we report the results of the best distillation portion $\alpha$ in each setting. Here, we provide results of other $\alpha$ settings. As shown in Figure~\ref{fig:alpha}, in high-IPC settings, the optimal distillation portion $\alpha$ is typically between 0.2 and 0.4. We recommend a small distillation portion $\alpha$ in high-IPC settings.

\section{CCFS with MTT}
In the main results, we use CDA to get $\cD_\textrm{distill}$. However, our curriculum selection framework is independent of the base dataset distillation method and can be applied to other dataset distillation methods. To verify the scalability of CCFS, we also provide results using MTT as the dataset distillation method. We compare them with SelMatch, which is also based on the MTT approach. We follow the same experimental setup as SelMatch to evaluate the synthetic datasets on ResNet-18. The results in Table~\ref{tab:mtt} demonstrate that CCFS with MTT still outperforms SelMatch across all high-IPC settings, showcasing its excellent scalability.

\section{More Experimental Results}
In the ablation study, we present the results of other combinations in the selection strategy on CIFAR-100 with IPC=50 and demonstrate that the simplest-misclassified strategy is the optimal combination. Here, we provide experimental results of more datasets and more IPC settings to further validate the effectiveness of the simplest-misclassified combination. As shown in Table~\ref{tab:strategy_cifar10},~\ref{tab:strategy_cifar100} and~\ref{tab:strategy_tiny}, the simplest-misclassified combination consistently outperforms others in all settings. This further validates the effectiveness of our coarse-to-fine selection strategy.

\begin{table}[h]
\setlength{\abovecaptionskip}{1mm}
\setlength{\belowcaptionskip}{-1mm}
\centering\scriptsize
\caption{CIFAR-10}
\label{tab:strategy_cifar10}
\begin{threeparttable}
\begin{tabular}{@{} c|cccccc @{}}
\toprule
\multirow{2}{*}{IPC} & \multicolumn{3}{c}{classified} & \multicolumn{3}{c}{misclassified} \\ \cmidrule(lr){2-4} \cmidrule(lr){5-7}
& random & hard & simple & random & hard & simple \\
\midrule
250 & 84.2 & 86.0 & 85.4 & 87.0 & 86.4 & \textbf{87.9} \\
500 & 89.8 & 90.5 & 90.8 & 91.8 & 91.6 & \textbf{92.5} \\
1000 & 91.5 & 91.8 & 91.9 & 92.6 & 92.2 & \textbf{93.2} \\
1500 & 92.4 & 93.0 & 92.9 & 93.2 & 92.9 & \textbf{93.8} \\
\bottomrule
\end{tabular}
\end{threeparttable}
\end{table}

\begin{table}[h]
\setlength{\abovecaptionskip}{1mm}
\setlength{\belowcaptionskip}{-2mm}
\centering\scriptsize
\caption{CIFAR-100}
\label{tab:strategy_cifar100}
\begin{threeparttable}
\begin{tabular}{@{} c|cccccc @{}}
\toprule
\multirow{2}{*}{IPC} & \multicolumn{3}{c}{classified} & \multicolumn{3}{c}{misclassified} \\ \cmidrule(lr){2-4} \cmidrule(lr){5-7}
& random & hard & simple & random & hard & simple \\
\midrule
25 & 59.2 & 52.5 & 60.5 & 62.9 & 51.6 & \textbf{65.3} \\
50 & 66.8 & 63.5 & 66.8 & 70.1 & 65.0 & \textbf{71.5} \\
100 & 70.7 & 69.1 & 70.4 & 72.0 & 71.0 & \textbf{73.0} \\
150 & 72.1 & 71.6 & 71.2 & 73.3 & 72.7 & \textbf{74.8} \\
\bottomrule
\end{tabular}
\end{threeparttable}
\end{table}

\begin{table}[h]
\setlength{\abovecaptionskip}{1mm}
\setlength{\belowcaptionskip}{-2mm}
\centering\scriptsize
\caption{Tiny-ImageNet}
\label{tab:strategy_tiny}
\begin{threeparttable}
\begin{tabular}{@{} c|cccccc @{}}
\toprule
\multirow{2}{*}{IPC} & \multicolumn{3}{c}{classified} & \multicolumn{3}{c}{misclassified} \\ \cmidrule(lr){2-4} \cmidrule(lr){5-7}
& random & hard & simple & random & hard & simple \\
\midrule
50 & 52.1 & 48.4 & 52.5 & 52.9 & 46.5 & \textbf{55.8} \\
100 & 58.1 & 56.4 & 57.7 & 58.2 & 54.9 & \textbf{60.2} \\
\bottomrule
\end{tabular}
\end{threeparttable}
\end{table}

\section{Visualization}
\begin{figure*}[htb!]
    \begin{center}
      \includegraphics[width=0.95\textwidth]{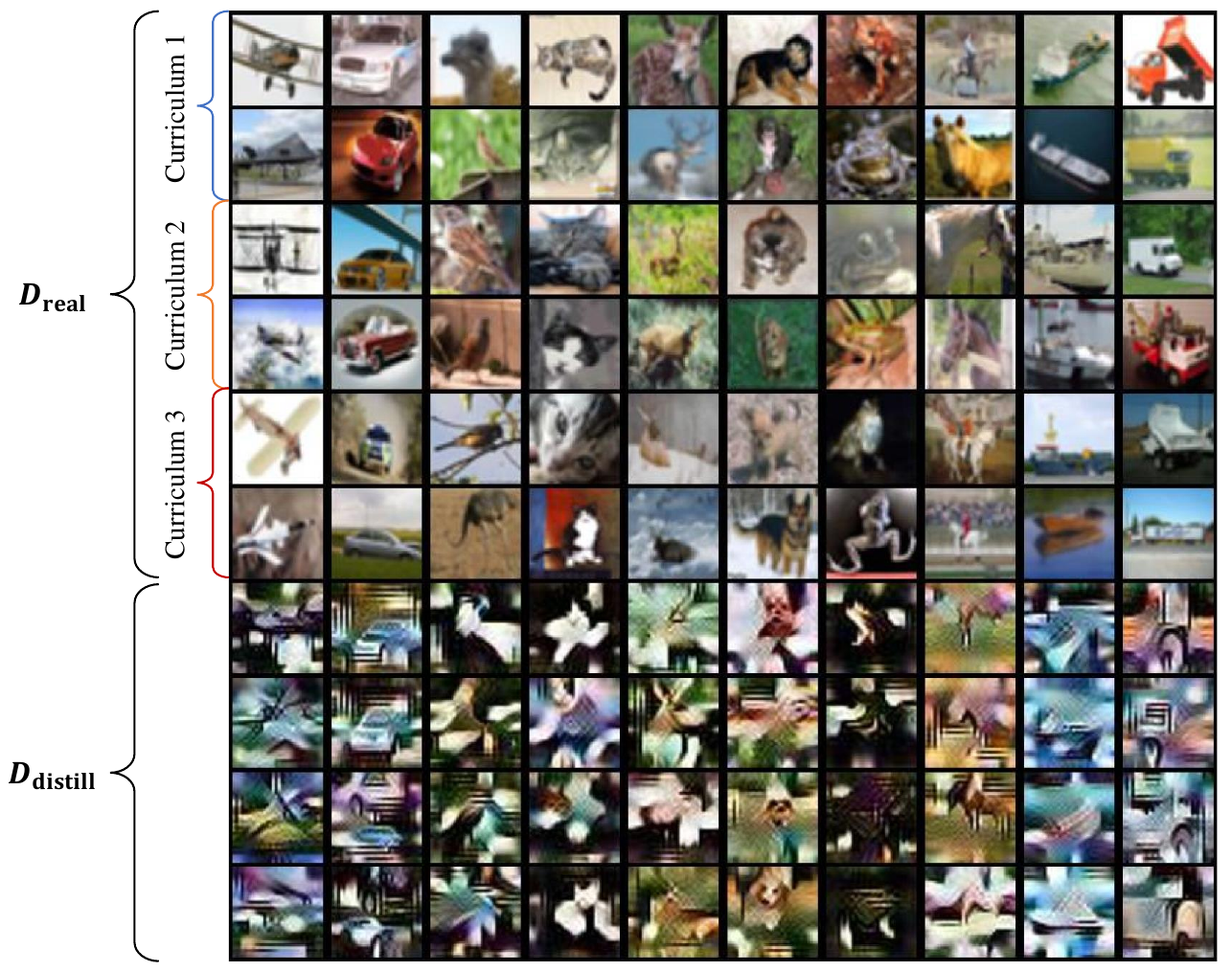}
    \end{center}
    \caption{Visualization of the synthetic dataset (CIFAR-10, IPC=250)}
    \label{fig:vis_cifar10_ipc250}
\end{figure*}

\begin{figure*}[htb!]
    \begin{center}
      \includegraphics[width=0.95\textwidth]{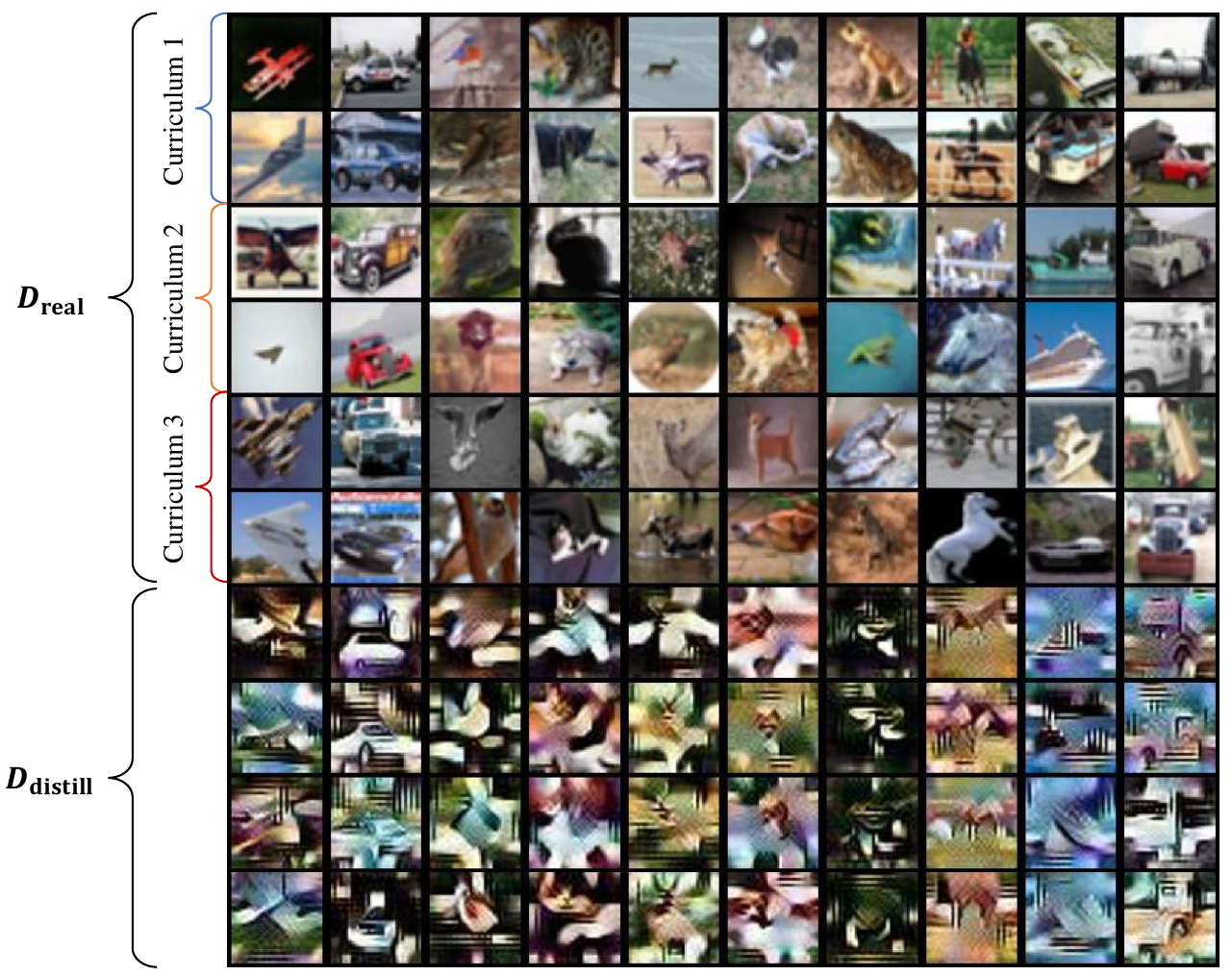}
    \end{center}
    \caption{Visualization of the synthetic dataset (CIFAR-10, IPC=1500)}
    \label{fig:vis_cifar10_ipc1500}
\end{figure*}

\begin{figure*}[htb!]
    \begin{center}
      \includegraphics[width=0.95\textwidth]{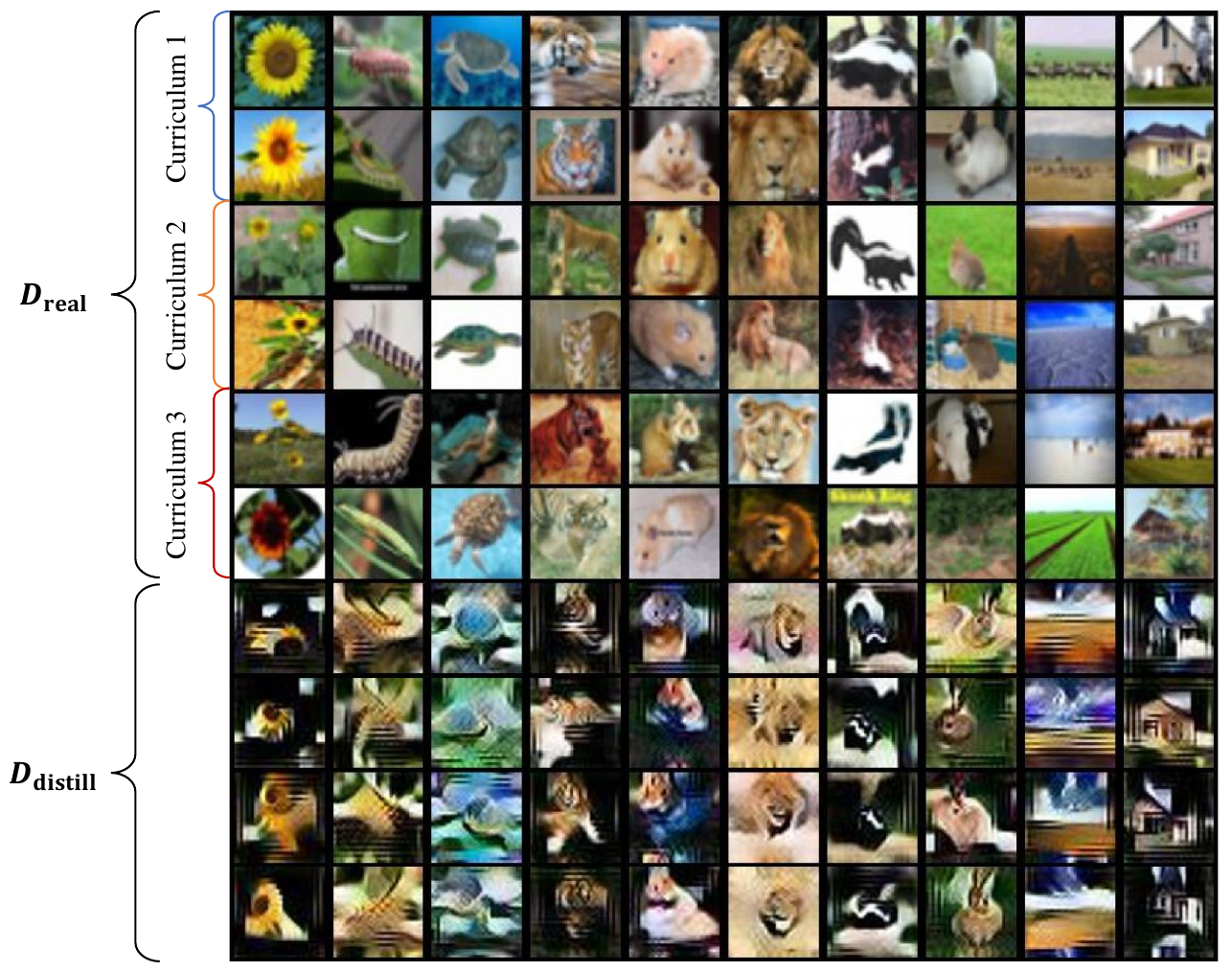}
    \end{center}
    \caption{Visualization of the synthetic dataset (CIFAR-100, IPC=25)}
    \label{fig:vis_cifar100_ipc25}
\end{figure*}

\begin{figure*}[htb!]
    \begin{center}
      \includegraphics[width=0.95\textwidth]{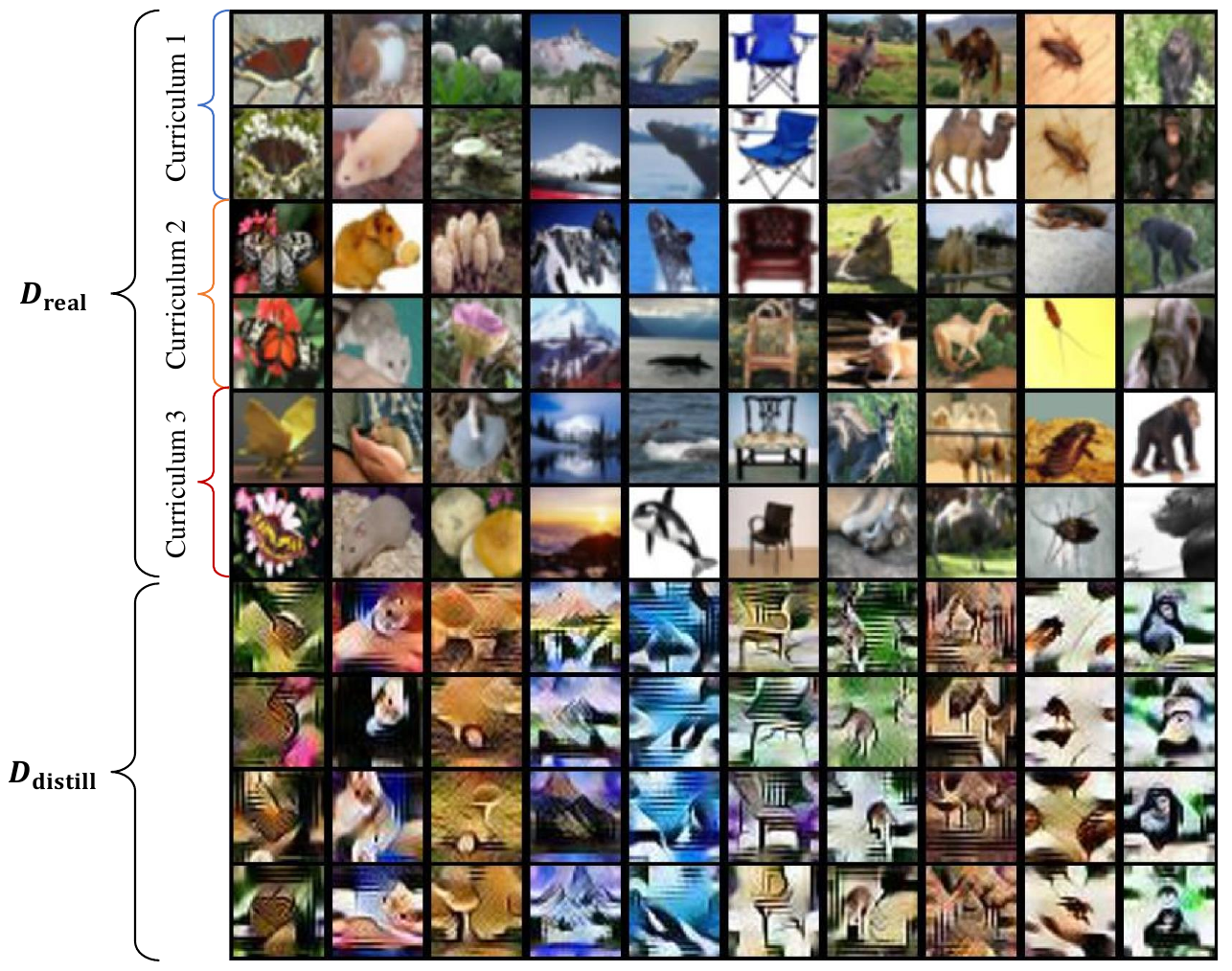}
    \end{center}
    \caption{Visualization of the synthetic dataset (CIFAR-100, IPC=150)}
    \label{fig:vis_cifar100_ipc150}
\end{figure*}

\begin{figure*}[htb!]
    \begin{center}
      \includegraphics[width=0.95\textwidth]{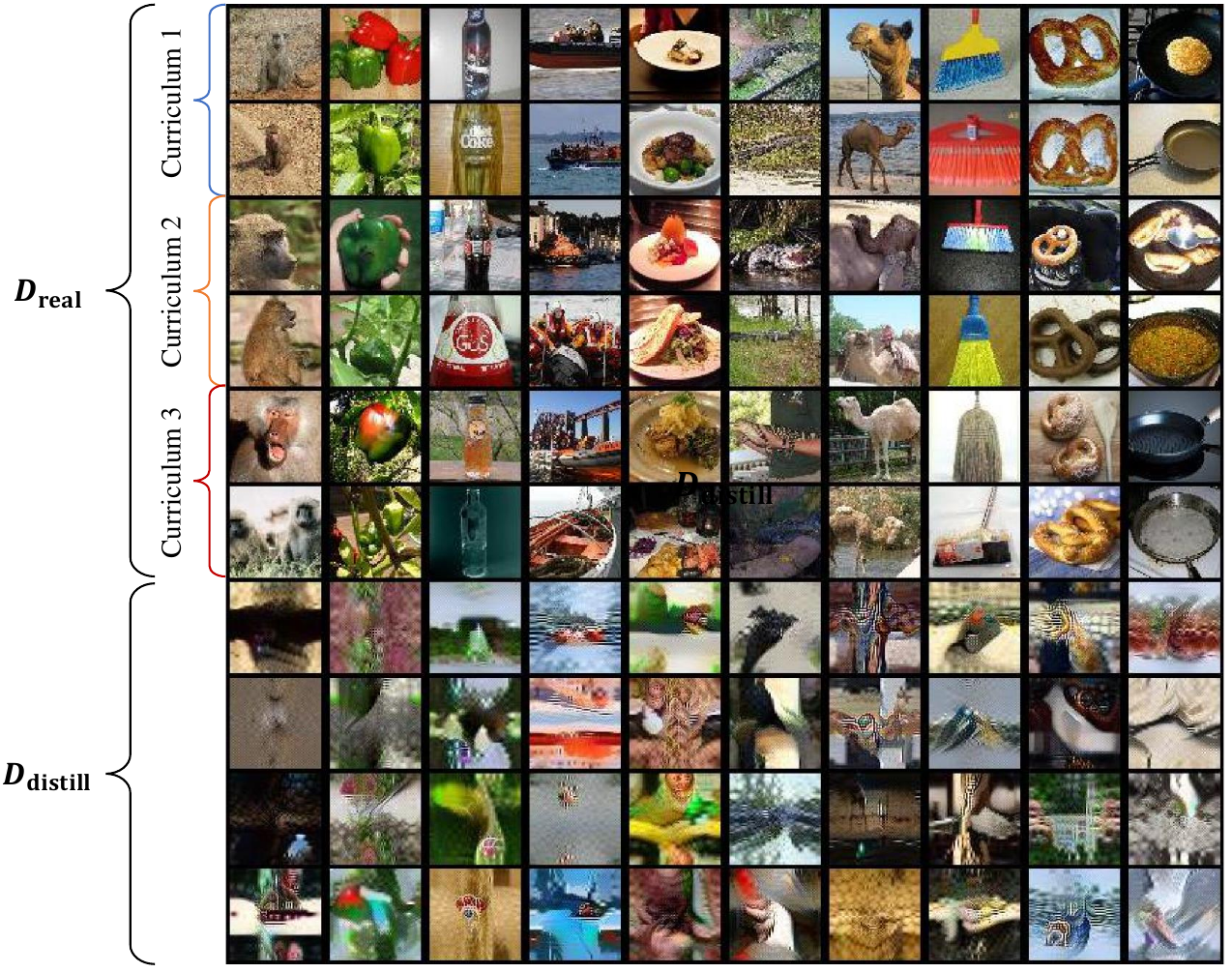}
    \end{center}
    \caption{Visualization of the synthetic dataset (Tiny-ImageNet, IPC=50)}
    \label{fig:vis_tiny_ipc50}
\end{figure*}

\begin{figure*}[htb!]
    \begin{center}
      \includegraphics[width=0.95\textwidth]{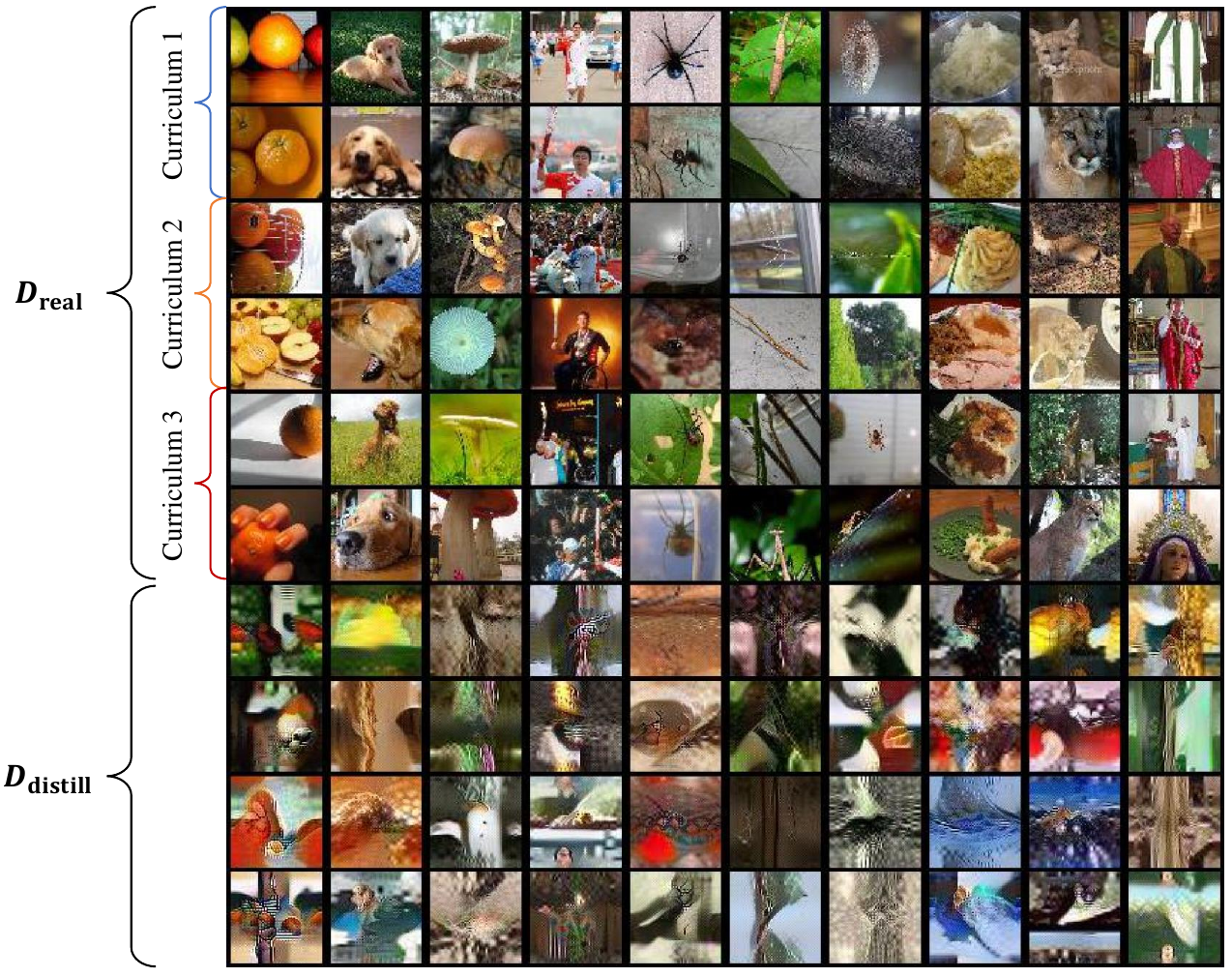}
    \end{center}
    \caption{Visualization of the synthetic dataset (Tiny-ImageNet, IPC=100)}
    \label{fig:vis_tiny_ipc100}
\end{figure*}

We present more visualizations of the synthetic datasets, including CIFAR-10 with IPC=250 (ratio=5\%) and 1500 (ratio=30\%) in Figure~\ref{fig:vis_cifar10_ipc250} and~\ref{fig:vis_cifar10_ipc1500}, resp., CIFAR-100 with IPC=25 (ratio=5\%) and 150 (ratio=30\%) in Figure~\ref{fig:vis_cifar100_ipc25} and~\ref{fig:vis_cifar100_ipc150}, resp., and Tiny-ImageNet with IPC=50 (ratio=10\%) and 100 (ratio=20\%) in Figure~\ref{fig:vis_tiny_ipc50} and~\ref{fig:vis_tiny_ipc100}, respectively. In each visualization, we show partial images from 10 classes in the dataset (corresponding to 10 columns). The first six rows denote the selected real images $\cD_\textrm{real}$, while the last four rows correspond to the distilled images $\cD_\textrm{distill}$. For $\cD_\textrm{real}$, we display two samples of median difficulty per class selected at each curriculum phase. The visualizations demonstrate the progressive difficulty of selected samples across curriculum phases and show that higher IPC settings tend to select more challenging samples than lower IPC settings within the same phase. 

